\documentclass[letterpaper, 10 pt, journal, twoside]{IEEEtran}  
\usepackage{amsmath,amsfonts}
\usepackage{algorithmic}
\usepackage{algorithm}
\usepackage{array}
\usepackage{textcomp}
\usepackage{url}
\usepackage{verbatim}
\usepackage{graphicx}
\usepackage{cite}
\usepackage{soul, color}
\usepackage{graphicx,adjustbox}

\usepackage{multirow} 
\usepackage{xcolor}   
\usepackage{siunitx}  
\usepackage{amssymb}  

\begin{document}

\title{Controlling Pose, Trajectory, and Stiffness of PCC-Modeled Soft Arms with Virtual Model Control}


\author{Lorenzo Vignoli$^{1,3}$, Ittirath Wiwatwanont$^{1}$, Kyle L. Walker$^{2}$, Fulvio Forni$^{3}$, and Josie Hughes$^{1}$%
\thanks{Manuscript received: Month, Day, Year; Revised Month, Day, Year; Accepted Month, Day, Year.}%
\thanks{$^{1}$L. Vignoli, I. Wiwatwanont, and J. Hughes are with the CREATE Lab, \'{E}cole Polytechnique F\'{e}d\'{e}rale de Lausanne (EPFL), Lausanne, Switzerland.}%
\thanks{$^{2}$K. L. Walker is with the Department of Mechanical and Aerospace Engineering, University of Liverpool, Liverpool, U.K.}%
\thanks{$^{3}$L. Vignoli and F. Forni are with the Department of Engineering, University of Cambridge, Cambridge, U.K.}%
}

\markboth{IEEE Robotics and Automation Letters. Preprint Version. Accepted Month, Year}{Vignoli \MakeLowercase{\textit{et al.}}: Controlling PCC-Modeled Soft Arms with Virtual Model Control}

\maketitle

\begin{abstract}
How a soft continuum arm interacts with its environment depends on the configuration of its whole body and on how it responds to contact. Piecewise constant curvature (PCC) models describe this configuration compactly, but are mostly used to control the tip through inverse kinematics, while full-configuration controllers require a complete target shape, and those that shape interaction act at the tip or within a plane. We propose Virtual Model Control (VMC) for soft arms, which attaches virtual springs, dampers, and repulsive fields along the arm and maps their forces onto the actuators through the Jacobians of a PCC model, without inverse kinematics or a model of the environment. Changing only the virtual elements controls the full configuration and trajectory of a tendon-driven arm, as well as the stiffness and a bounded force at its tip.
With hand-designed virtual mechanisms, the arm exploits, regulates, or avoids contact, also under teleoperation.
In contact, the stiffness stays bounded by the arm's compliance, keeping interaction safe. Virtual Model Control thus offers an interpretable way to control the full pose of a soft robot arm and its response to contact.
\end{abstract}


\section{Introduction}

\IEEEPARstart{S}{oft} continuum arms can deform along their entire body length, enabling configurations where they can exploit their body for environment interactions~\cite{santinaModelBasedControl2023}.
This configuration space and their inherent compliance make them well suited to tasks in unstructured, confined, or human-centered environments, where distributed and safe full-body interaction with the environment is advantageous or required.
These tasks require controlling the pose and behavior of the full body.
Reduced-order models, most notably piecewise constant curvature (PCC)~\cite{dellasantinaImprovedStateParametrization2020}, represent the full backbone of soft robots compactly.
Yet PCC is mostly used to control the tip through inverse kinematics, rather than the full configuration or the stiffness of the body~\cite{santinaModelBasedControl2023}.
We therefore need methods that use such full-body representations to intuitively control the pose of the full body and how it responds to contact.

\begin{figure*}[!t]
    \centering
    \includegraphics[width=0.93\linewidth]{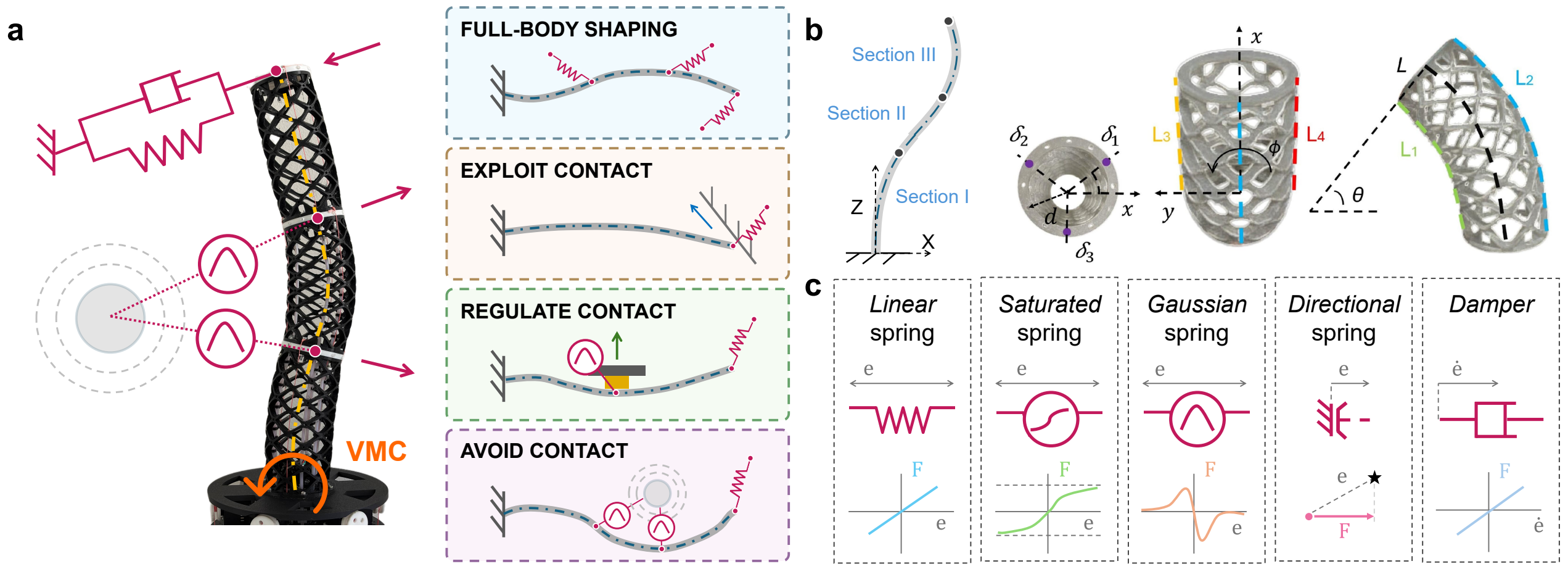}
    \caption{Virtual Model Control (VMC) of a soft continuum arm. (a)~Virtual elements connected at the tip and along the body form a single virtual mechanism; changing only the elements switches the arm between shaping its body and exploiting, regulating, or avoiding contact. (b)~PCC kinematics of the tendon-driven arm. (c)~Virtual elements and their force laws, Eq.~\eqref{eq:elements}.}
    \label{fig:1}
\end{figure*}

PCC and other reduced-order representations, such as affine~\cite{stellaPiecewiseAffineCurvature2023} and polynomial~\cite{santinaControlOrientedModeling2020} curvature models, enable task-space control of the tip through inverse kinematics (IK)~\cite{kapadiaTaskspaceControlExtensible2011}, learned planning~\cite{chenVersatileNeuralNetwork2025}, or adaptive schemes~\cite{veroneseFeedbackErrorLearning2024}.
Approaches that do control the full configuration regulate the PCC curvatures through model-based feedback~\cite{dellasantinaDynamicControlSoft2018, dellasantinaModelbasedDynamicFeedback2020}, shape the potential energy so that a desired configuration becomes a stable equilibrium~\cite{francoEnergyshapingControlSoft2021, borjaEnergybasedShapeRegulation2022}, or solve closed-loop IK while reasoning about the entire continuous shape~\cite{veilInfiniteDimensionalClosedLoopInverse2026}.
These approaches control the body toward a target configuration, requiring the user to either explicitly define the full pose or rely on IK, which must resolve the redundancy of the arm.

A second challenge is controlling stiffness and, more generally, the response to contact (Table~\ref{tab:comparison}).
Model-based dynamic feedback has enabled surface following with planar soft robots~\cite{dellasantinaModelbasedDynamicFeedback2020}, the Cartesian stiffness of the end effector has been prescribed by co-optimizing posture and physical stiffness~\cite{stellaPrescribingCartesianStiffness2023}, and Cartesian impedance control has rendered direction-dependent compliance at the tip of a planar robot~\cite{stolzleGuidingSoftRobots2024}.
Control Barrier Functions (CBFs) have bounded tip contact forces against a known deformable environment~\cite{dicksonSafeAutonomousEnvironmental2025}, enforced contact-force constraints in planar simulations~\cite{wongContactAwareSafetySoft2025}, and kept the whole body clear of obstacles~\cite{wongClosedFormCLFCBF2026}.
In these works, interaction is shaped only at the tip or within a plane, and CBFs rely on accurate robot or environment models~\cite{caiSafetyControlHyperredundant2026} and treat contact as a constraint to be bounded or avoided, rather than a capability to exploit.

We propose to combine a full-body representation of soft arms, such as PCC, with Virtual Model Control (VMC) to intuitively control the configuration of the full body and its contact stiffness.
VMC attaches virtual mechanical elements, such as springs, dampers, and potential fields, to points of the robot and maps their forces into actuator commands through the kinematic Jacobians, so that the robot behaves as if the elements were physically present~\cite{larbyOptimalVirtualModel2026}.
Attached anywhere along the backbone, these elements provide physically interpretable control of the robot's 3D configuration, motion, and interaction stiffness, without IK or replanning.
Because each element is passive, the closed loop remains passive, and thus stable in contact with passive environments~\cite{hatanakaPassivitybasedControlRobots2015}, while complementing rather than canceling the robot's own elasticity.
We validate the approach on a three-section tendon-driven trimmed-helicoid arm~\cite{guanTrimmedHelicoidsArchitectured2023} (Fig.~\ref{fig:1}), where virtual elements control its configuration, trajectory, tip force and stiffness, and contact within a single framework.

Although VMC has been used in locomotion~\cite{prattVirtualModelControl2001}, surgical robots~\cite{larbyGeneralizedApproachImpedance2023}, and compliant reaching~\cite{zhangVirtualModelControl2024}, to the best of our knowledge, this letter presents its first extension to soft continuum arms.
Unlike prior work on soft arms, VMC regulates the full-body configuration in 3D, with virtual stiffness at any body point (Table~\ref{tab:comparison}). This letter contributes:
\begin{itemize}
    \item The development of VMC for soft continuum arms, for the first time, where virtual elements connected along a PCC model specify both the goals and the response to contact across the whole body without IK or replanning, together with an approach to designing such virtual mechanisms.
    \item Characterization experiments on whole-body shaping, end-effector control with a moving obstacle, directional position control, and the force and stiffness rendered at the tip, including their limits.
    \item Demonstrations of environmental interaction in which the arm exploits, regulates, or avoids contact by changing only the virtual elements: surface following, body-contact regulation, obstacle avoidance, and shared control in a teleoperated handover.
\end{itemize}


The remainder of this letter presents the model and controller (Section~\ref{sec:methods}), the experimental setup (Section~\ref{sec:experimental_setup}), the results on position and force control (Section~\ref{sec:results}) and on task execution (Section~\ref{sec:interaction}), and conclusions (Section~\ref{sec:conclusion}). The implementation is openly available.\footnote{\url{https://github.com/CREATE-Lab-EPFL/SoftVMC}}

\begin{table*}[!t]
\centering
\caption{Comparison of Controllers for Soft Continuum Arms That Address Interaction}
\label{tab:comparison}
\footnotesize
\setlength{\tabcolsep}{4pt}
\renewcommand{\arraystretch}{1.1}
\begin{tabular*}{\textwidth}{@{\extracolsep{\fill}}l c l l l l@{}}
\hline
\textbf{Approach} & \textbf{Dim.} & \textbf{Objectives at} & \textbf{Stiffness set at} & \textbf{Contact is} & \textbf{Solved online}\\
\hline
Planning with a learned model~\cite{chenVersatileNeuralNetwork2025} & 3D & tip & --- & avoided & optimization\\
Model-based dynamic feedback~\cite{dellasantinaModelbasedDynamicFeedback2020} & 2D & tip & --- & exploited (surface following) & none\\
Posture--stiffness co-optimization~\cite{stellaPrescribingCartesianStiffness2023} & 3D & tip & tip, via variable-stiffness segments & prescribed (tip stiffness) & optimization\\
Cartesian impedance~\cite{stolzleGuidingSoftRobots2024} & 2D & tip & tip & prescribed (tip stiffness) & none\\
CBF safety filter~\cite{dicksonSafeAutonomousEnvironmental2025} & 2D & tip & --- & bounded at the tip & QP\\
High-order CBF and CLF~\cite{wongContactAwareSafetySoft2025} & 2D & tip and body & --- & bounded along the body & QP\\
Closed-form CLF--CBF~\cite{wongClosedFormCLFCBF2026} & 3D & tip and body & --- & avoided & none\\
\hline
\textbf{VMC with a PCC model (ours)} & \textbf{3D} & \textbf{tip and body} & \textbf{virtual, any body point} & \textbf{avoided, exploited, or regulated} & \textbf{none}\\
\hline
\end{tabular*}\\[2pt]
\parbox{\textwidth}{\footnotesize CBF/CLF: Control Barrier/Lyapunov Function; QP: Quadratic program; ---: not addressed.}
\end{table*}

\section{Methods}\label{sec:methods}

VMC requires the kinematics of the robot, mapping motor coordinates into points along the robot body, and virtual passive elements, whose forces are mapped back onto the motors. Since the controller uses only the model Jacobians, and never their inverse, we use PCC as the kinematic model, a low-order description of the full configuration of a soft arm. In this section, we first formulate PCC for a tendon-driven soft arm, then describe VMC, and finally how to design a virtual mechanism for a task.

\subsection{Piecewise Constant Curvature on Tendon-Driven Soft Arms}\label{subsec:pcc_soft}

We formulate PCC for our test platform, a tendon-driven, three-section trimmed-helicoid soft arm~\cite{guanTrimmedHelicoidsArchitectured2023} (Fig.~\ref{fig:2}a).
Each section is modeled as an arc of constant curvature~\cite{jonesKinematicsMultisectionContinuum2006}. Rather than the bending angle $\theta$ and bending direction $\phi$ (Fig.~\ref{fig:1}b), which become singular when a section is straight, we describe each section with the parameters $[\Delta_x, \Delta_y, L]^{T}$ of~\cite{dellasantinaImprovedStateParametrization2020}, where $\Delta_x = \theta d \cos\phi = (L_2 - L_1)/2$ and $\Delta_y = \theta d \sin\phi = (L_4 - L_3)/2$ are half the length differences between opposite fibers at distance $d$ from the central axis ($L_1$--$L_4$ in Fig.~\ref{fig:1}b), and $L$ is the length of the central axis. The three tendons of a section run at angles $\delta_i$ around this axis, and their lengths depend linearly on these parameters:
\begin{equation}
l_i = L - \Delta_x \cos\delta_i - \Delta_y \sin\delta_i = L_0 + r\,q_i,
\label{eq:tendons}
\end{equation}
where $L_0$ is the rest length and $q_i$ the angle of the motor that winds tendon $i$ on a spool of radius $r$. Inverting Eq.~\eqref{eq:tendons} section by section gives the configuration of the whole arm from the nine motor encoders alone. With $\Delta_{xy} = (\Delta_x^2 + \Delta_y^2)^{1/2} = \theta d$, the point at a fraction $s \in [0,1]$ of a section lies at:
\begin{equation}
\mathbf{t}(s) = \frac{d\,L}{\Delta_{xy}^2}
\begin{bmatrix}
\Delta_x \left(1 - \cos\frac{s\Delta_{xy}}{d}\right) \\
\Delta_y \left(1 - \cos\frac{s\Delta_{xy}}{d}\right) \\
\Delta_{xy} \sin\frac{s\Delta_{xy}}{d}
\end{bmatrix}
\label{eq:pcc}
\end{equation}
in the frame at its base, with the corresponding rotation given in~\cite{dellasantinaImprovedStateParametrization2020}. Concatenating the three sections gives the position $\mathbf{x}(s,\mathbf{q})$ of any point along the backbone, where $s \in [0,1]$ is now the curvilinear coordinate along the whole arm, and its Jacobian $\mathbf{J}(s,\mathbf{q}) = \partial \mathbf{x}/\partial \mathbf{q} \in \mathbb{R}^{3\times 9}$, both derived symbolically and evaluated at every control cycle. The controller only needs a Jacobian that points the virtual forces in the right direction, which PCC provides analytically and at low cost. Other tendon routings or actuation schemes only change Eq.~\eqref{eq:tendons}.

\subsection{Virtual Model Control for Continuum Manipulators}\label{subsec:vmc_continuum}

In VMC, virtual mechanical elements are connected to points of the robot, and their forces are mapped onto the actuators through virtual work~\cite{larbyOptimalVirtualModel2026, prattVirtualModelControl2001}. Rather than tracking a trajectory, the controller specifies a virtual mechanism. Each element $k$ has two terminals: one is connected to the robot at the body point $\mathbf{x}_k = \mathbf{x}(s_k,\mathbf{q})$, with $s_k$ its curvilinear coordinate, the other to a point $\mathbf{x}_{\text{ref},k}$ in the workspace, its anchor. The element exerts on the robot a force $\mathbf{f}_k$ that depends on the offset between its terminals and on its rate, and the motor torques are
\begin{equation}
\boldsymbol{\tau} = \sum_k \mathbf{J}_k^{T}\, \mathbf{f}_k(\mathbf{e}_k, \dot{\mathbf{e}}_k) + \boldsymbol{\tau}_g, \quad \mathbf{e}_k = \mathbf{x}_{\text{ref},k} - \mathbf{x}_k,
\label{eq:vmc}
\end{equation}
where $\mathbf{J}_k = \mathbf{J}(s_k,\mathbf{q})$ and $\boldsymbol{\tau}_g$ compensates gravity. Each element adds its own term, any number of elements can share the nine motors, and their terminals and parameters can be changed while the arm moves~\cite{zhangVirtualModelControl2024, larbyGeneralizedApproachImpedance2023}, so reaching a goal and responding to contact come from one virtual mechanism.

Although any virtual element can be included, we use five classes (Fig.~\ref{fig:1}c). Dropping the index $k$, their forces are
\begin{equation}
\begin{aligned}
&\text{Linear spring:} && \mathbf{f} = K\mathbf{e},\\
&\text{Saturated spring:} && \mathbf{f} = F_{\max}\tanh\left(K\mathbf{e}/F_{\max}\right),\\
&\text{Gaussian spring:} && \mathbf{f} = -A \exp\left(-\|\mathbf{e}\|^2/2\sigma^2\right)\mathbf{e},\\
&\text{Directional spring:} && \mathbf{f} = K\,\mathbf{n}\mathbf{n}^{T}\mathbf{e},\\
&\text{Damper:} && \mathbf{f} = D\,\dot{\mathbf{e}},
\end{aligned}
\label{eq:elements}
\end{equation}
where $K$, $F_{\max}$, $A$, $\sigma$, and $D$ are positive constants, $\mathbf{n}$ is a unit vector, and $\tanh$ is applied componentwise.

The linear spring pulls the body point toward its anchor with stiffness $K$, the ratio of force to offset. The saturated spring behaves as a linear spring for $|K e_i| \ll F_{\max}$, but each component of its force saturates at $F_{\max}$ however far the anchor is. The Gaussian spring instead pushes the body point away from its anchor, placed in a region to avoid: its force is 
that of a negative stiffness $-A$ for small offsets, peaks at $A\sigma e^{-1/2}$ for $\|\mathbf{e}\| = \sigma$, and decays to about $5\%$ of this peak at $3\sigma$, so it acts only near its anchor. The directional spring is a linear spring whose anchor slides freely on a plane with unit normal $\mathbf{n}$, so it pulls only along $\mathbf{n}$. The damper has a fixed anchor, so $\dot{\mathbf{e}} = -\dot{\mathbf{x}}$, and dissipates energy at the rate $D\|\dot{\mathbf{x}}\|^2$. Each spring force derives from a potential, $\mathbf{f} = \partial U/\partial \mathbf{e}$, the energy stored in the spring, which is non-negative and bounded for finite offsets:
\begin{equation}
\begin{aligned}
&U = \tfrac{1}{2}K\|\mathbf{e}\|^2, && U = \tfrac{F_{\max}^2}{K}\textstyle\sum_i \ln\cosh\left(K e_i/F_{\max}\right),\\
&U = A\sigma^2 e^{-\|\mathbf{e}\|^2/2\sigma^2}, && U = \tfrac{1}{2}K(\mathbf{n}^{T}\mathbf{e})^2,
\end{aligned}
\label{eq:storage}
\end{equation}
for the linear, saturated, Gaussian, and directional springs, respectively. The Gaussian potential and force are bounded by $A\sigma^2$ and $A\sigma e^{-1/2}$, so adding this element cannot inject unbounded energy or force.

The stability of VMC rests on a basic result of mechanics: springs and dampers can store and dissipate energy but never create it, and passive systems connected together remain passive, hence stable. With fixed anchors and parameters, the controller is such a mechanism: as Eq.~\eqref{eq:vmc} maps each force through the Jacobian of its own body point, the power the elements deliver to the arm is the rate $-\tfrac{d}{dt}\sum_k U_k$ at which the springs release energy minus the power dissipated by the dampers, so Eq.~\eqref{eq:vmc} derives from a total potential plus dissipation. The arm, itself a viscoelastic body, is passive too, so the closed loop is passive~\cite{albu-schafferUnifiedPassivitybasedControl2007, hatanakaPassivitybasedControlRobots2015} and stable in contact with any passive environment.
Due to the passivity property, stability is independent of the 
model parameters. 
This matters for soft arms, whose dynamics are uncertain: stability holds whatever the accuracy of the PCC model, which only affects where the virtual forces act.

In practice, moving any anchor point $\mathbf{x}_{\text{ref},k}$ 
to achieve a specific motion or changing the stiffness of the virtual elements can inject energy and break passivity~\cite{kronanderStabilityConsiderationsVariable2016}. In our experiments, anchors were moved slowly or were stepped and held.
This guarantees that the (bounded) amount of energy injected into the system is dissipated by the virtual dampers, the arm, and its tendon transmission, restoring passivity. More advanced treatments can be found in \cite{ferragutiTankbasedApproachImpedance2013}.

\subsection{Designing Controllers Using VMC}\label{sec:design}

As shown in Fig.~\ref{fig:1}a, the design of a controller corresponds to the design of a virtual mechanism, and each task adds its own elements to it. The structure of the mechanism, which elements are used and where their terminals are connected, sets the direction of each force. Its tuning, the parameters $K$, $F_{\max}$, $A$, $\sigma$, and $D$, sets their magnitude. 
In this work, terminals are connected at the ends of the first two sections and at the tip. A spring at the tip pulls it toward its anchor, while elements at the other points shape the rest of the body. Pulling a point of a PCC arm with a spring is like pulling a flexible rod at that point: the sections between the base and the pulled point bend toward the anchor until the arm's elasticity balances the virtual force, so the shape is specified only at the task-relevant points, not as a full target configuration. The other elements set the response to contact: a directional spring leaves the tip free within a plane, a saturated spring bounds the force, and a Gaussian spring keeps a point of the body away from a region. Moving the anchors over time generates a trajectory without a time-parametrized reference, and elements can be switched on or off or driven by different sources, including human operators.

\section{Experimental Setup}\label{sec:experimental_setup}

\subsection{Robot Design and Fabrication}

\begin{figure}[!t]
    \centering
    \includegraphics[width=0.95\linewidth]{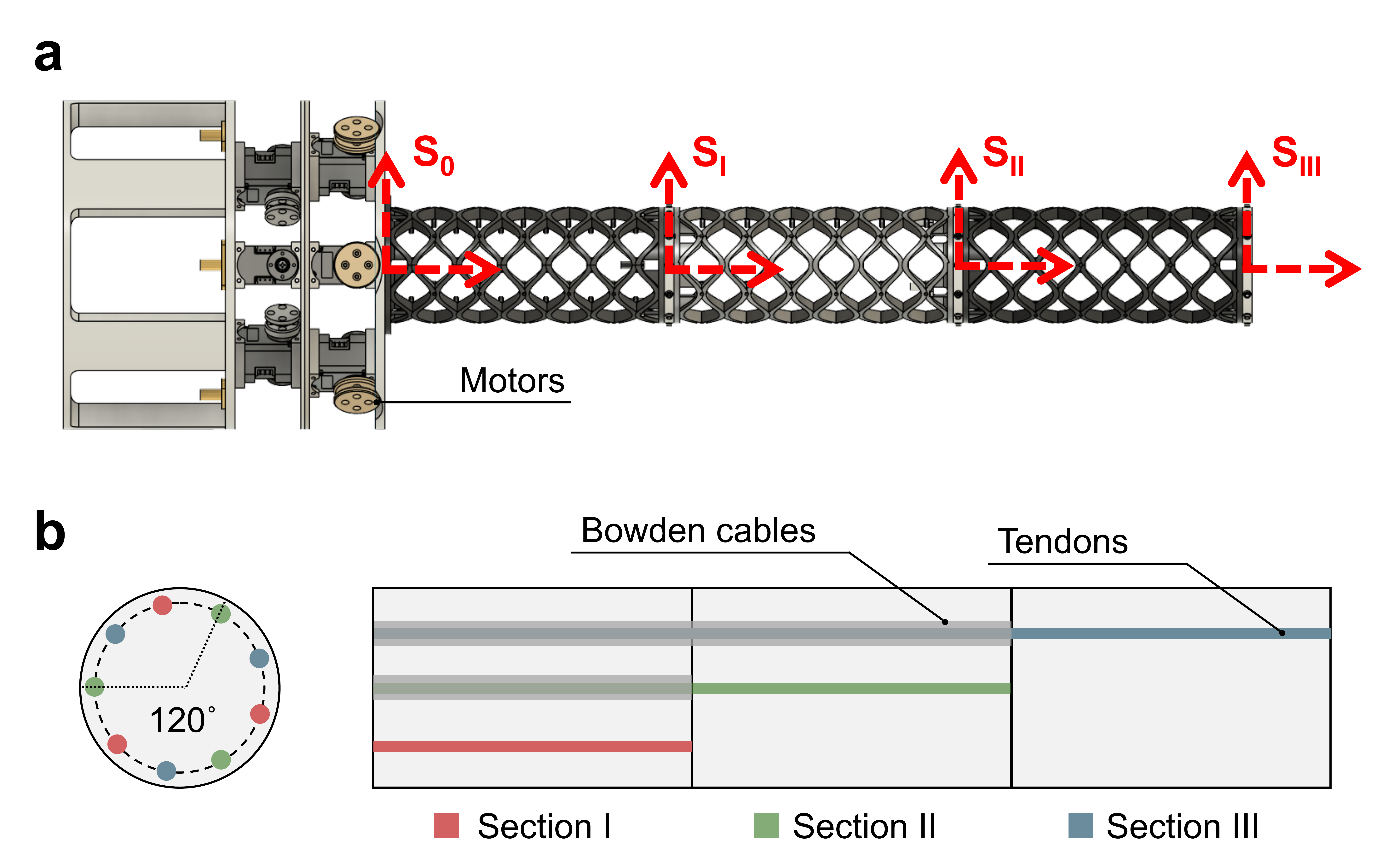}
    \caption{Experimental setup. (a)~Trimmed-helicoid arm and actuation unit, with the PCC section frames S$_0$--S$_\mathrm{III}$. (b)~Tendon routing; sections rotated by $0^\circ$, $60^\circ$, and $150^\circ$ w.r.t.\ the base.}
    \label{fig:2}
\end{figure}

Our three-section arm is formed from 3D-printed thermoplastic polyurethane (TPU 95A) trimmed-helicoid lattice sections with an outer diameter of $60~\text{mm}$ and a nominal length of $45~\text{cm}$~\cite{guanTrimmedHelicoidsArchitectured2023}, which bend and compress. Nine Dynamixel \mbox{XL330-M288-T} motors in the base, three per section, actuate the tendons through spools of $3~\text{mm}$ radius. The tendons are spaced $120^\circ$ apart, and those of Sections~II and~III pass through the proximal sections inside Bowden cables, isolating the bending moment (Fig.~\ref{fig:2}b). The motors operate in current control mode, with the output torque assumed proportional to the current, so the torques of Eq.~\eqref{eq:vmc} can be commanded directly. The controller runs at $330~\text{Hz}$ in ROS~2. Gravity is compensated by mapping a lumped mass at the center of each section through the Jacobians, and a weak virtual spring on the motor angles prevents tendon slack. The arm was upright for Figs.~\ref{fig:3}--\ref{fig:8}, horizontal for Fig.~\ref{fig:9}, and hanging for Figs.~\ref{fig:10}--\ref{fig:11}.

\subsection{Force Measurement and Position Estimation}

We measure interaction forces with a load cell (HX711 amplifier, read by an Arduino), logged with the controller state. To characterize the tip stiffness (Figs.~\ref{fig:6}--\ref{fig:7}), a tether, slack at rest, connects the tip to the load cell, which records the shear force along it. On the body (Fig.~\ref{fig:9}), the load cell records the shear force on a vertical rod next to the arm. Positions are estimated from motor encoders via the PCC model, except in Fig.~\ref{fig:8}, where the tip was tracked in video.

\subsection{Graphical User Interface (GUI)}

A graphical user interface (GUI) displays the estimated PCC backbone and lets the user change the virtual mechanism while the arm moves, by dragging anchors, tuning $K$, $A$, and $D$, or switching elements on and off; dragging the tip anchor turns it into a teleoperation input (Fig.~\ref{fig:11}). Scripts can publish the same commands, and an anchor trajectory recorded from the GUI can be replayed with other elements.

\section{Experimental Results:\\ Position and Force Control}\label{sec:results}


\begin{figure}[!t]
    \centering
    \includegraphics[width=0.874\linewidth]{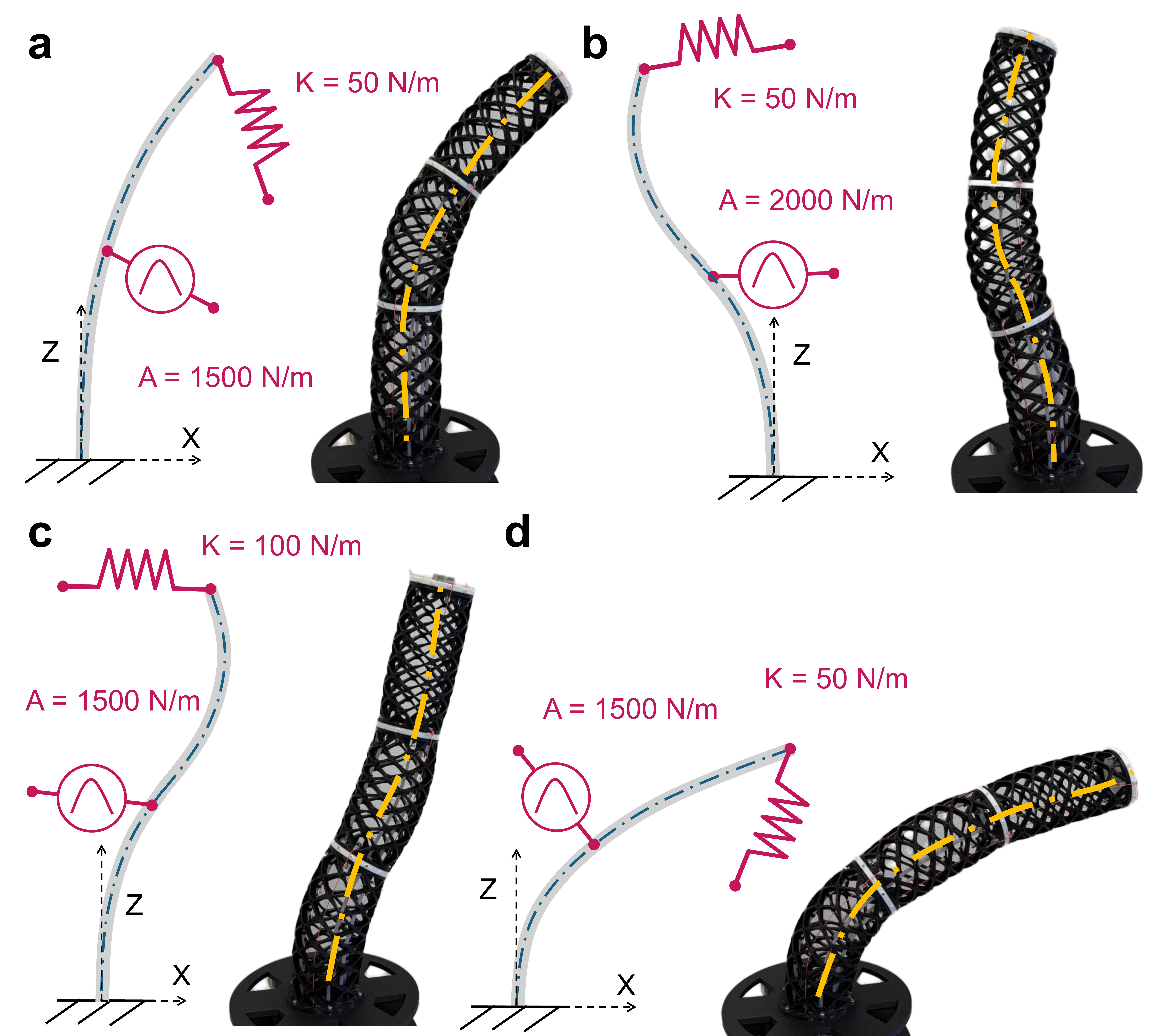}
    \caption{Full-body shaping with a linear tip spring and a Gaussian spring at the end of the first section. (a)--(d)~Schematic of the virtual mechanism (left) and photograph with the estimated backbone overlaid (right).}
    \label{fig:3}
\end{figure}
\begin{figure}[!t]
    \centering
    \includegraphics[width=0.874\linewidth]{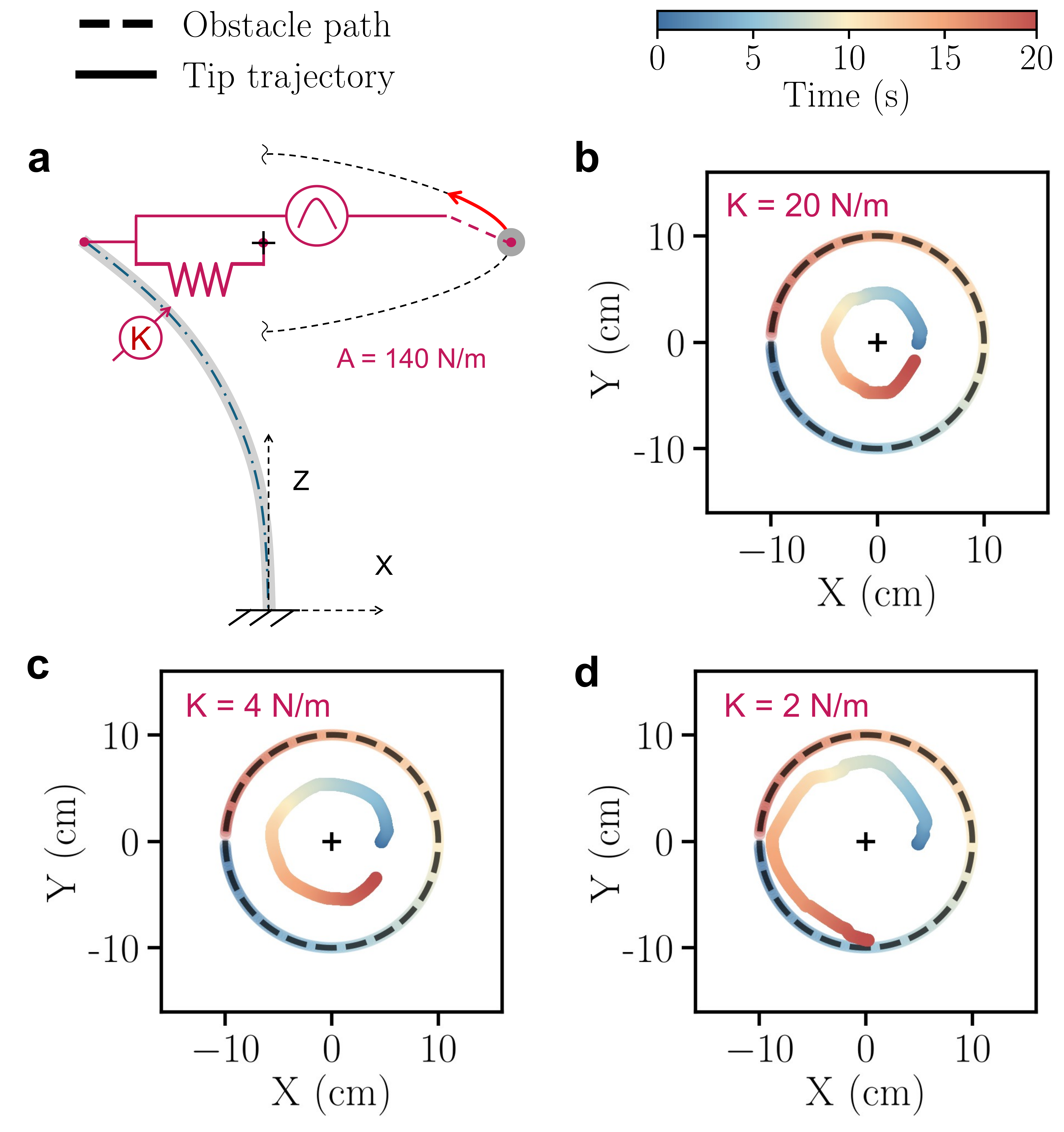}
    \caption{End-effector control with a moving obstacle. (a)~A linear spring pulls the tip toward the center ($+$) of a $10~\text{cm}$ radius circle, while the anchor of a Gaussian spring on the tip, the obstacle, moves around the circle (red arrow). (b)--(d)~Tip trajectories for decreasing stiffness $K$.}
    \label{fig:4}
\end{figure}

\subsection{Full-Body Pose Control}

We want to control the shape of the whole arm, not only the position of its tip. A continuum arm can reach the same tip position with multiple body configurations, which IK-based controllers must resolve explicitly~\cite{kapadiaTaskspaceControlExtensible2011}; in VMC, the body shape instead emerges from the placement of the virtual elements. As shown in Fig.~\ref{fig:3}, two elements are enough: a linear tip spring ($K = 50$ or $100~\text{N/m}$) and a Gaussian spring at the end of the robot's Section~I ($A = 1500\text{--}2000~\text{N/m}$, $\sigma = 6~\text{cm}$), anchored beside the body, which pushes Section~I away from its anchor. Their balance with the arm's elasticity yields a single bend (C-shaped, Fig.~\ref{fig:3}a, d) or, when Section~I is pushed far enough to the side opposite to the tip anchor, a double bend (S-shaped, Fig.~\ref{fig:3}b, c). Unlike a second linear spring, the Gaussian spring pushes locally without pinning the body point to an anchor. Virtual elements placed along the backbone thus allow intuitive control of the full pose.

\subsection{End-Effector Control with a Moving Obstacle}\label{sec:trajectory}

Next, we want to hold the tip at a point while an obstacle moves around it: a trade-off between staying centered and keeping clear of the obstacle. As shown in Fig.~\ref{fig:4}a, a linear spring of stiffness $K$ anchors the tip at the center of a $10~\text{cm}$ radius circle, while the anchor of a Gaussian spring on the tip ($A = 140~\text{N/m}$, $\sigma = 6~\text{cm}$), the obstacle, travels around the circle in $20~\text{s}$. The tip orbits opposite the obstacle, at a radius set by the balance of attraction and repulsion of the two springs (Fig.~\ref{fig:4}b--d), so $K$ sets the trade-off: reducing $K$ increases the radius, as long as the orbit settles. The orbit radius settles at about $4.6~\text{cm}$ for $K = 20~\text{N/m}$ and $5.5~\text{cm}$ for $K = 4~\text{N/m}$ (Fig.~\ref{fig:4}b, c). At $K = 2~\text{N/m}$, the spring no longer balances the repulsion, which, like a push on a stick, is unstable sideways: the orbit does not settle, its radius grows to about $9~\text{cm}$, and the tip no longer stays opposite the obstacle (Fig.~\ref{fig:4}d).

\subsection{Directional Position Control}

\begin{figure}[!t]
    \centering
    \includegraphics[width=0.95\linewidth]{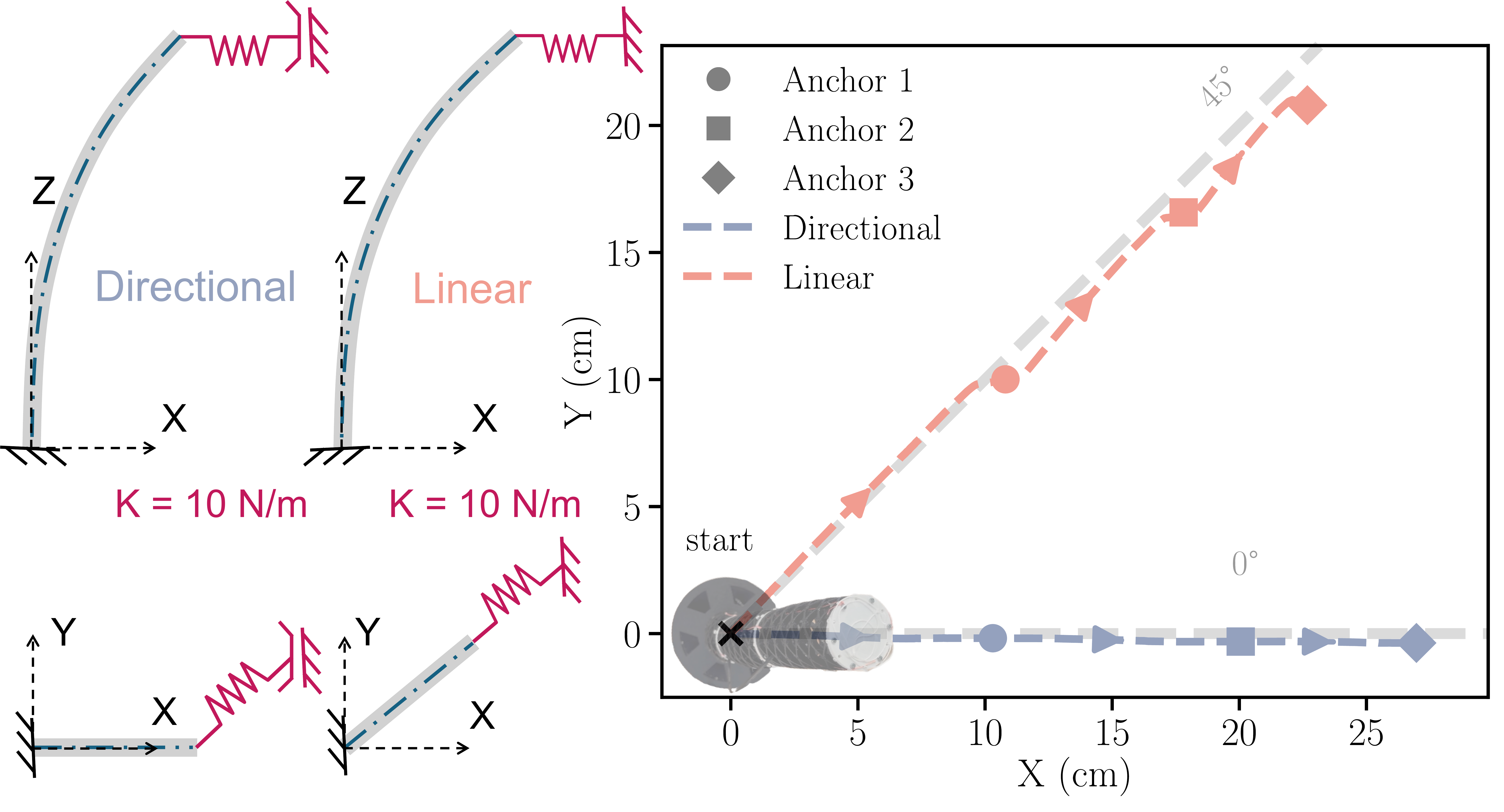}
    \caption{Directional versus linear spring on the tip ($K = 10~\text{N/m}$), with the anchor stepped along the $45^\circ$ diagonal. Left: side ($xz$) and top ($xy$) schematics. Right: tip trajectories in the $xy$ plane, with markers $1~\text{s}$ after each anchor step.}
    \label{fig:5}
\end{figure}

\begin{figure}[!t]
    \centering
    \includegraphics[width=0.93\linewidth]{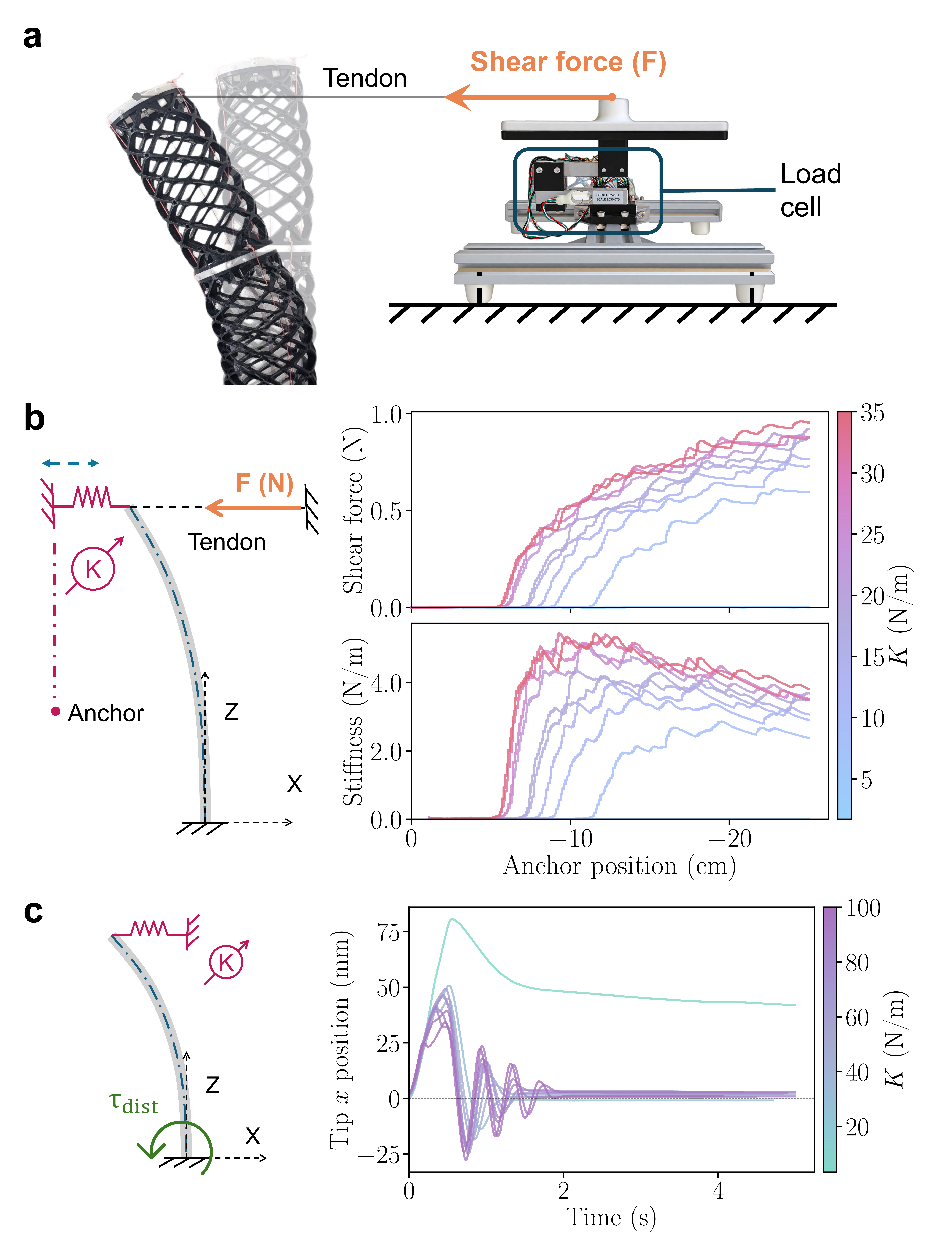}
    \caption{Tip force and stiffness. (a)~Setup: a tether joins the tip to a load cell. (b)~Shear force and secant stiffness as the anchor moves outward at $1~\text{cm/s}$, colored by the virtual stiffness $K$. (c)~Tip response to a commanded torque pulse $\tau_{\text{dist}}$ ($-0.3~\text{N}{\cdot}\text{m}$, $0.5~\text{s}$) on the $0^\circ$ tendon of Section~I and the $\pm 60^\circ$ tendons of Section~II, without virtual damping.}
    \label{fig:6}
\end{figure}

Fig.~\ref{fig:5} compares a linear and a directional spring on the tip, both with $K = 10~\text{N/m}$. Here we want to constrain the tip motion to one direction, $\mathbf{n} = (1, 0, 0)$, and to control its position along it with a given stiffness, as needed to guide a tool along a slot or to press on a surface while sliding across it. For both springs, the anchor is stepped through three points on the $45^\circ$ diagonal, $(20, 20, 40)$, $(40, 40, 40)$, and $(60, 60, 30)~\text{cm}$. The linear spring pulls the tip along the diagonal, reaching $(22.8, 21.0)~\text{cm}$, whereas the directional spring moves it only along $\mathbf{n}$, advancing $27.1~\text{cm}$ along $x$ while staying within $0.4~\text{cm}$ of the $x$ axis (Fig.~\ref{fig:5}, right), although the arm has no physical guide and the constraint exists only in the controller. Since the directional spring exerts no force within its plane, virtual elements regulate the direction of the force, not only its magnitude.

\begin{figure}[!t]
    \centering
    \includegraphics[width=0.95\linewidth]{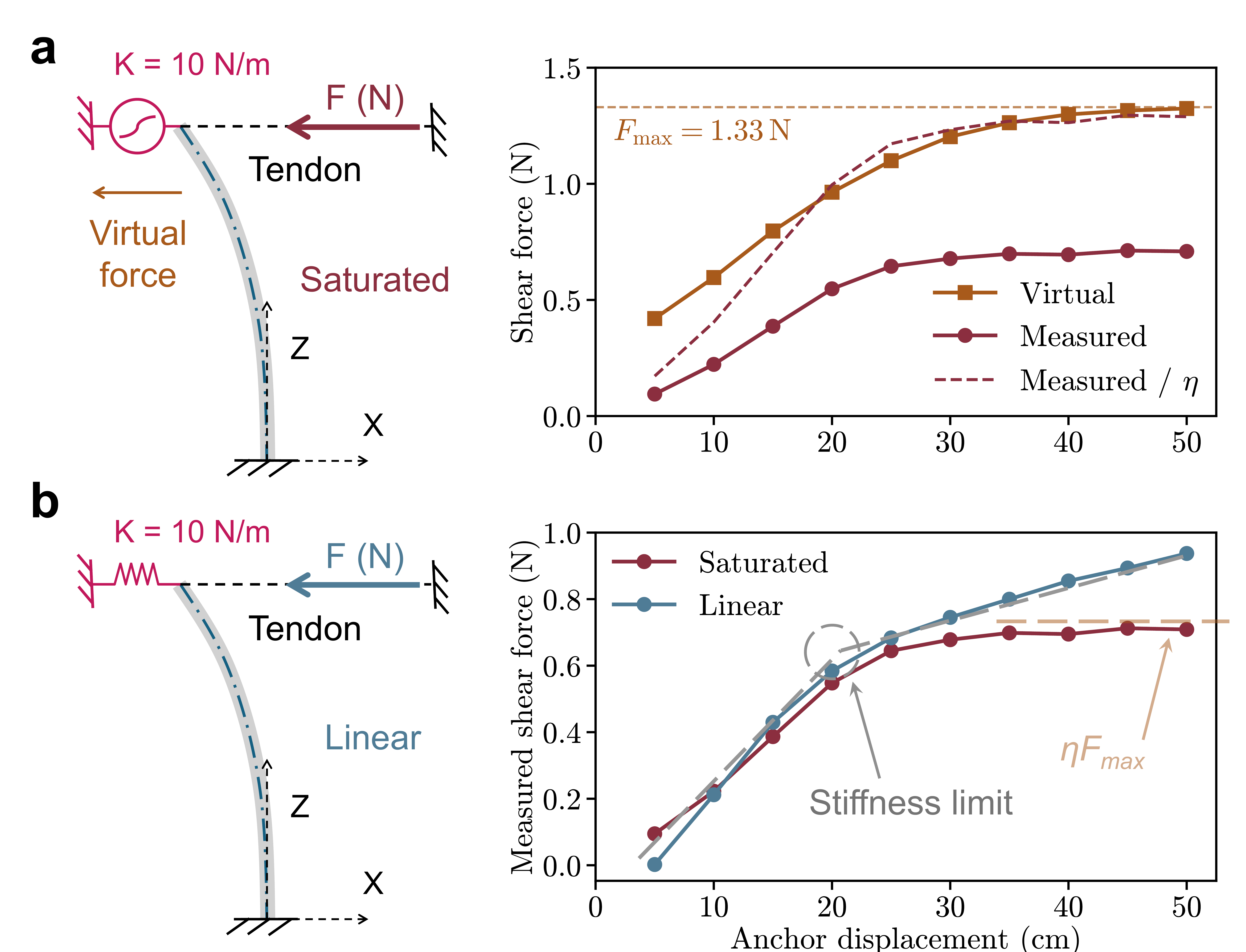}
    \caption{Saturated versus linear spring on the tether, for anchor jumps of $5\text{--}50~\text{cm}$ held for $3~\text{s}$. (a)~Saturated spring: virtual and measured force, and measured force divided by the transmission efficiency $\eta = 0.55$. (b)~Measured forces, with the bound $\eta F_{\max}$ and the stiffness limit of the linear spring highlighted.}
    \label{fig:7}
\end{figure}
\begin{figure*}[!t]
    \centering
    \includegraphics[width=0.9\linewidth]{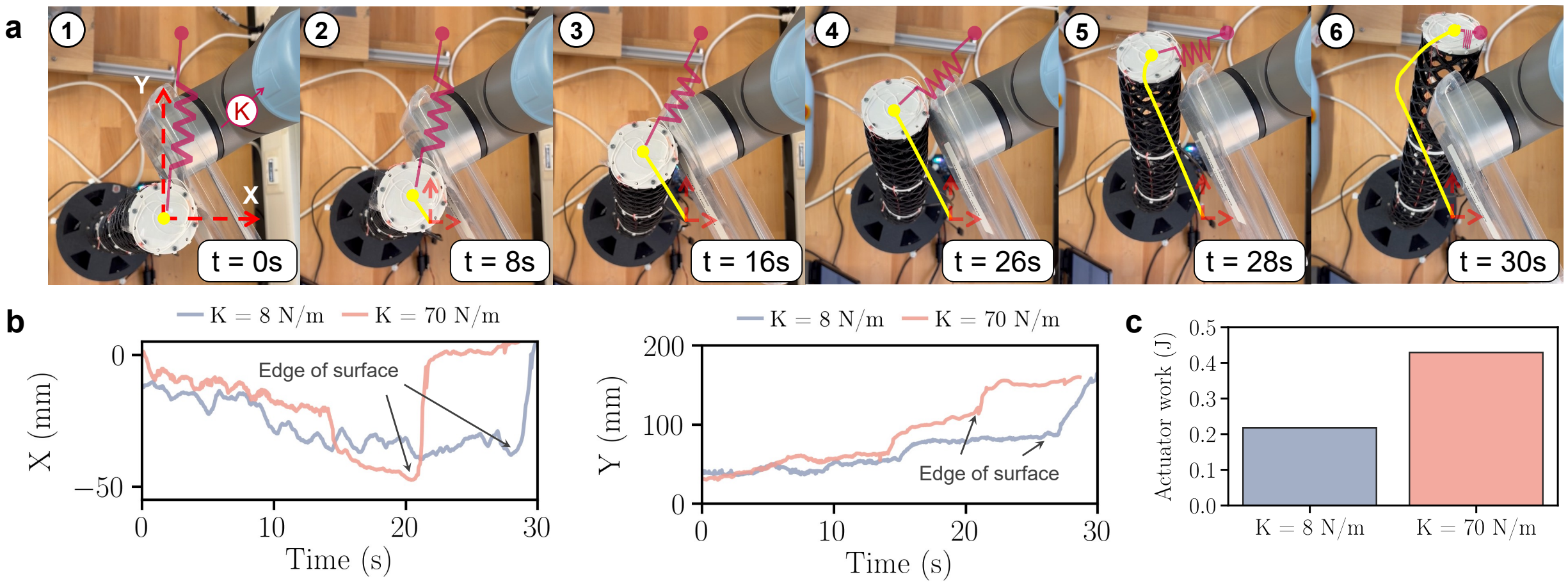}
    \caption{Surface following with a linear tip spring anchored behind an unmodeled slanted surface. (a)~Snapshots for $K = 8~\text{N/m}$, with the tip path in yellow. (b)~Tracked tip $x$ and $y$ positions (camera frame) for a soft and a stiff spring; arrows: tip clears the edge. (c)~Actuator work over the first $45~\text{s}$.}
    \label{fig:8}
\end{figure*}
\subsection{Stiffness and Force Control}\label{sec:stiffness}

We now want to know which stiffness and force VMC renders at the tip. As shown in Fig.~\ref{fig:6}a, we tether the tip to a load cell and move the anchor of a linear spring away from it at $1~\text{cm/s}$ for different $K$. A force is measured once the tether is taut, which takes a larger anchor displacement for low $K$. The secant stiffness $|F|/|x_{\text{ref}}|$, the ratio of force to anchor displacement, increases with $K$ up to $K \approx 20~\text{N/m}$ and then levels off at about $5~\text{N/m}$ (Fig.~\ref{fig:6}b). This limit comes from the series compliance of the arm and transmission, which dominates for stiff virtual springs.

We also want to know how $K$ affects the dynamic behavior, in particular the rejection of disturbances such as unexpected contacts. As shown in Fig.~\ref{fig:6}c, we apply a torque pulse $\tau_{\text{dist}}$ of $-0.3~\text{N}{\cdot}\text{m}$ for $0.5~\text{s}$ on the tendons of Sections~I and~II, with the tip held by a virtual spring without virtual damping. Increasing $K$ reduces the peak deflection, and the tip returns to rest within about $2~\text{s}$, with oscillations damped by the hardware. At the lowest stiffness, friction prevents the tip from returning, while at high stiffness the peak deflection levels off. VMC thus modulates the tip stiffness within the physical compliance limits of the arm.

Stiffness controls force only indirectly: in the static regime, when the anchor is out of reach, e.g., beyond the workspace or behind an obstacle, a spring applies a constant force. To characterize this force, we compare a saturated spring ($K = 10~\text{N/m}$, $F_{\max} = 1.33~\text{N}$) with a linear spring of equal stiffness (Fig.~\ref{fig:7}). Both produce similar forces for small gaps with the anchor, but for larger ones the measured force of the saturated spring levels off at about $0.71~\text{N}$, while that of the linear spring reaches $0.94~\text{N}$ at $50~\text{cm}$ (Fig.~\ref{fig:7}b). Divided by a constant transmission efficiency $\eta = 0.55$, close to the measured-to-commanded ratio at saturation, the measured force follows the commanded one for gaps above $20~\text{cm}$ (Fig.~\ref{fig:7}a). The saturated spring thus bounds the tip force near $\eta F_{\max} \approx 0.73~\text{N}$.

\begin{figure*}[!t]
    \centering
    \includegraphics[width=0.95\linewidth]{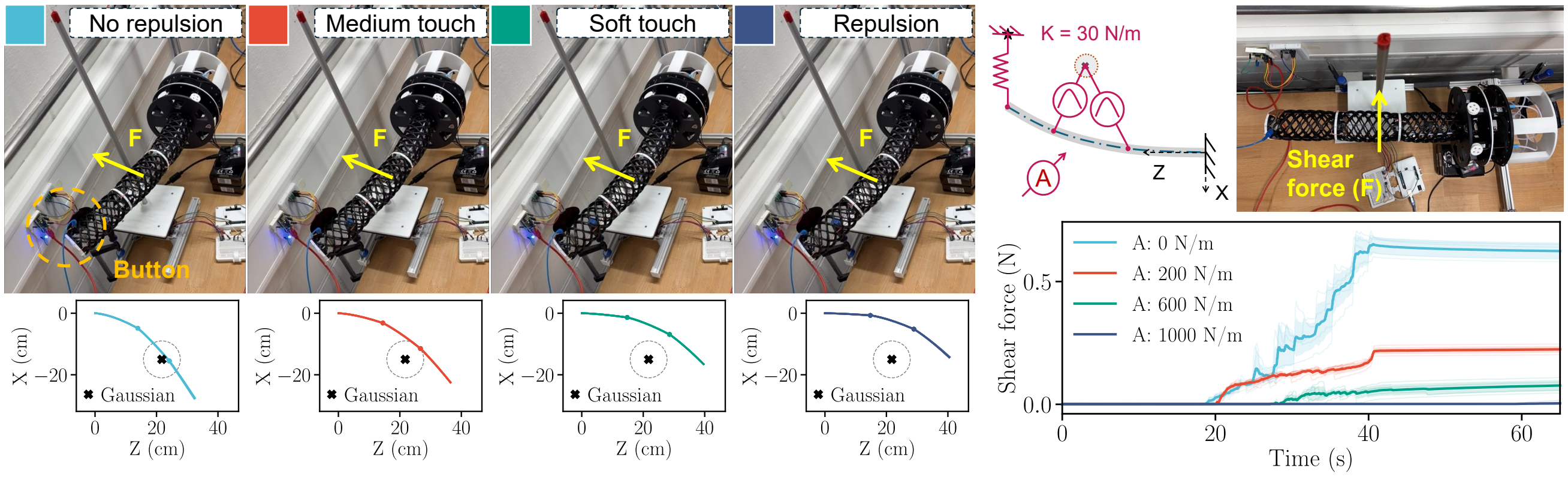}
    \caption{Contact regulation at a body point. A tip spring pulls the tip toward a button, while two Gaussian springs anchored on a rod push the body away from it with strength $A$ (cross: their anchor). Left: photographs, with the shear force $F$ that the body exerts on the rod (yellow), and arm configuration in the $xz$ plane. Right: setup and measured $F$ ($\text{mean}~\pm~\text{SD}$, five runs per condition).}
    \label{fig:9}
\end{figure*}
\section{Experimental Results: Task Execution}\label{sec:interaction}

\begin{figure*}[!t]
    \centering
    \includegraphics[width=0.95\linewidth]{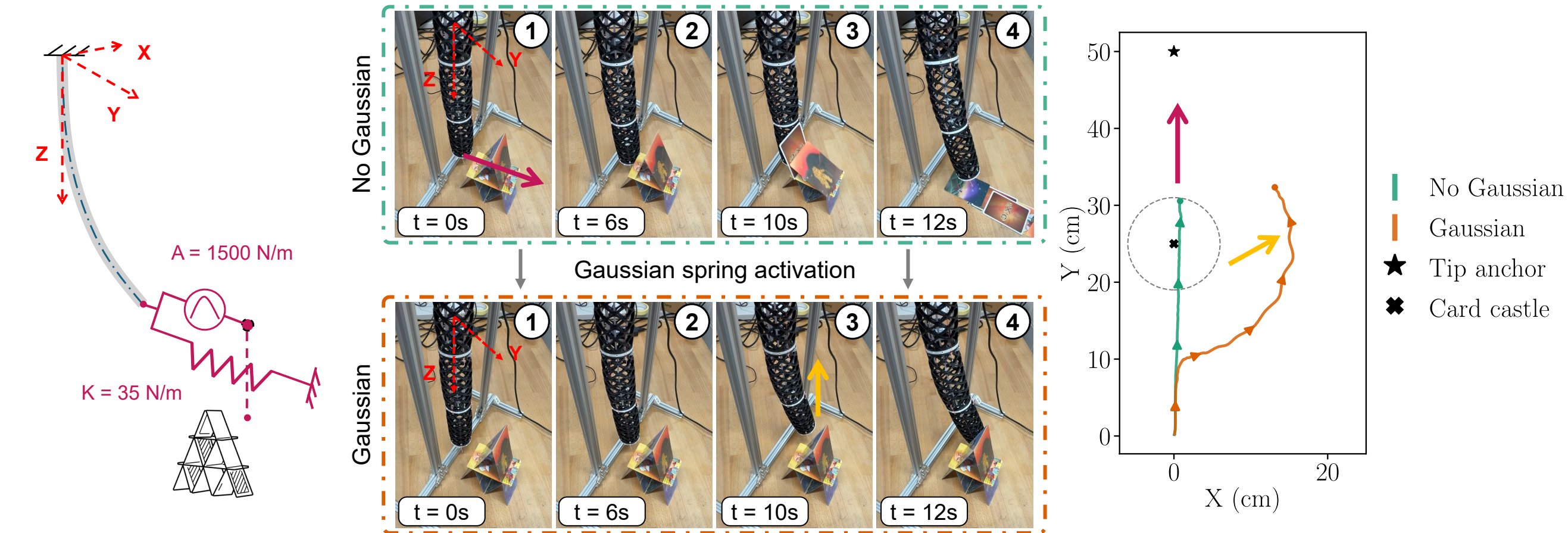}
    \caption{Obstacle avoidance. The anchor of a tip spring moves at $5~\text{cm/s}$ past a card castle toward an unreachable point, without (top) and with (bottom) a Gaussian spring on the tip anchored on the castle. Without it, the tip knocks the castle down; with it, the castle stands. Right: tip trajectories in the $xy$ plane.}
    \label{fig:10}
\end{figure*}

\begin{figure*}[!t]
    \centering
    \includegraphics[width=0.95\linewidth]{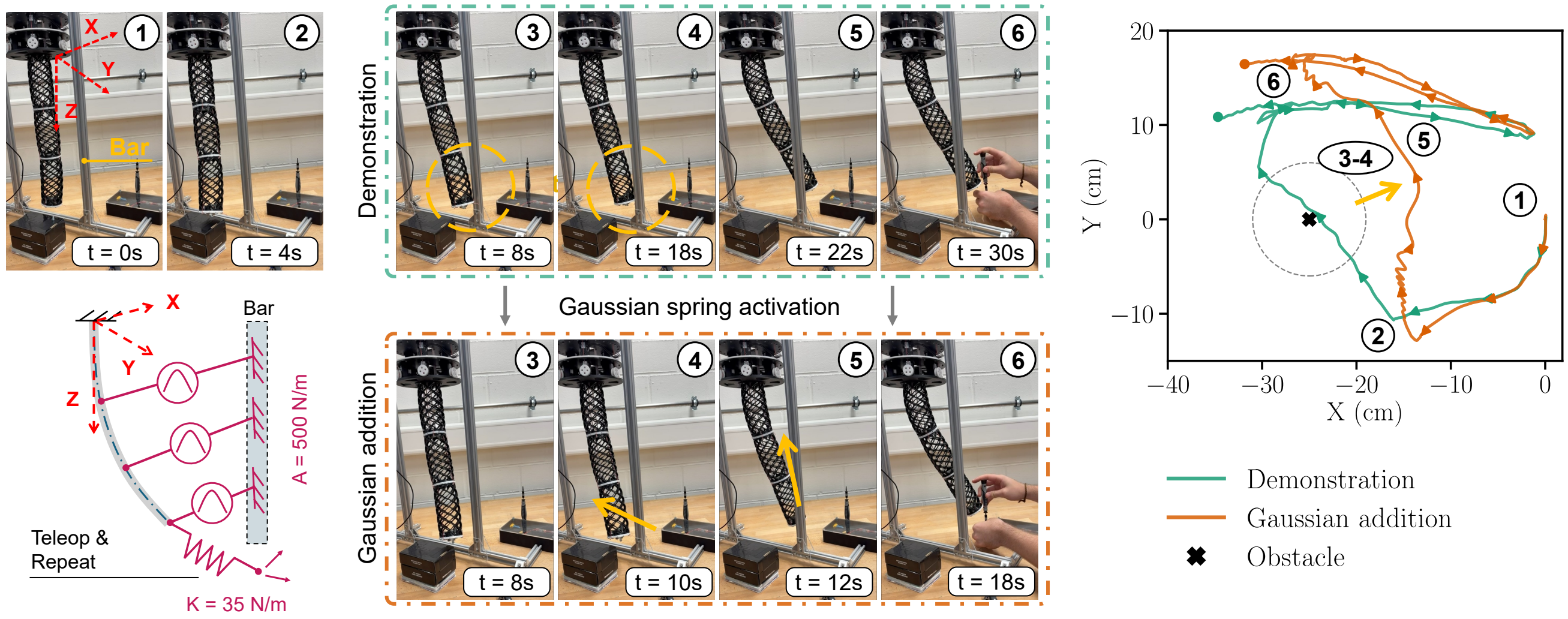}
    \caption{Teleoperated handover of a screw past a bar; panels 1--2 are common to both runs. Top: demonstration, with the operator moving the tip anchor from the GUI. Bottom: the same anchor trajectory replayed with three Gaussian springs anchored on the bar. Right: tip trajectories in the $xy$ plane.}
    \label{fig:11}
\end{figure*}


\subsection{Moving Along an Obstacle}
Here we want the tip to reach a target behind an unmodeled obstacle. A spring specifies a direction of attraction rather than a path, so a tip spring whose anchor moves toward a point behind the obstacle drags the tip along it. As shown in Fig.~\ref{fig:8}a, the anchor of a linear tip spring moves at $1~\text{cm/s}$ toward a point behind a slanted rigid surface held by a robot arm: the tip reaches the surface (2), slides along it (3--4), clears its edge (5), and continues toward the anchor (6). We repeat this experiment with a soft ($K = 8~\text{N/m}$) and a stiff ($K = 70~\text{N/m}$) spring, and Fig.~\ref{fig:8}b shows the same sequence for both, with abrupt changes when the tip clears the edge. The stiffer spring clears the edge earlier (about $21~\text{s}$ vs.\ $29~\text{s}$), but needs nearly twice the actuator work over the first $45~\text{s}$ ($0.43~\text{J}$ vs.\ $0.22~\text{J}$, Fig.~\ref{fig:8}c), computed as $E = \int \sum_{i=1}^{9} |\tau_i \dot{q}_i|\, dt$. The arm thus exploits contact to reach a goal without a prescribed path or environment model, with stiffness trading speed for effort.

\subsection{Contact Force Regulation Along the Robot Body}
In cluttered spaces, the body may touch the environment while the tip performs a task, so the contact force must be regulated. Here, the tip must press a button that lights an LED while the body passes a vertical rod mounted on a load cell. As shown in Fig.~\ref{fig:9}, we design a mechanism with a linear tip spring ($K = 30~\text{N/m}$) that attracts the tip toward the button, and two Gaussian springs at the ends of robot Sections~I and~II, anchored on the rod ($\sigma = 6~\text{cm}$), which push the body away from it, with their strength $A$ setting how hard the body presses. The tip anchor moves at $2~\text{cm/s}$ past the button, so that the tip presses it, and $A$ is set to $0$, $200$, $600$, or $1000~\text{N/m}$, with five runs each. As $A$ increases, the body moves away from the rod (Fig.~\ref{fig:9}, left), and the shear force decreases from $0.62 \pm 0.04~\text{N}$ without repulsion to $0.22 \pm 0.02~\text{N}$ at $A = 200~\text{N/m}$ and $0.08 \pm 0.03~\text{N}$ at $600~\text{N/m}$ (mean $\pm$ SD, Fig.~\ref{fig:9}, right), while at $1000~\text{N/m}$ contact is eliminated. With low variation across repetitions and the button pressed in all runs, body contact is regulated repeatably without compromising the tip task.

\subsection{Obstacle Avoidance}
When contact must be avoided, the same Gaussian element provides obstacle avoidance. Here, the tip must pass a card castle without knocking it down: as shown in Fig.~\ref{fig:10}, the anchor of a tip spring ($K = 35~\text{N/m}$) moves at $5~\text{cm/s}$ past the castle toward an unreachable point. Without repulsion, the tip takes the shortest path and knocks the castle down. With a Gaussian spring on the tip anchored on the castle ($A = 1500~\text{N/m}$, $\sigma = 6~\text{cm}$), the tip is deflected, passes at least $14.6~\text{cm}$ from the castle center, and leaves it standing. As the tip is closer to the castle than to the anchor, a small lateral displacement tilts the repulsion more than the attraction, so the equilibrium in front of the castle is unstable sideways, as for the push in Section~\ref{sec:trajectory}, and the net force guides the tip around it. Adding one element thus lets the tip avoid the obstacle without path planning.

\subsection{Shared Control via Teleoperation}
Finally, we want an operator to command only the tip while the rest of the body avoids an obstacle by itself, in shared control between the operator and the virtual mechanism. As shown in Fig.~\ref{fig:11} (bottom left), the mechanism has two parts: a tip spring ($K = 35~\text{N/m}$) whose anchor the operator moves from the GUI, and three fixed Gaussian springs at the ends of the sections, anchored on a vertical bar ($A = 500~\text{N/m}$, $\sigma = 6~\text{cm}$), which keep the whole body away from the bar. In a first run without the Gaussian springs (demonstration), the operator moves the anchor to pick up a screw with a magnet and hand it to a person, passing the bar. We then replay the same anchor trajectory with the Gaussian springs; as the torques are computed live, the two runs differ only in the virtual mechanism. In the demonstration, the arm stays against the bar from $8$ to $18~\text{s}$ (3--4) and completes the handover at $30~\text{s}$ (6); with the Gaussian springs, the arm clears the bar (3--5) and completes the handover at $18~\text{s}$ (6), following the anchor instead of being held back. Whole-body obstacle avoidance is thus obtained without additional operator commands or trajectory planning. Across all four tasks, intuitively designed virtual mechanisms let the arm exploit, regulate, or avoid contact without environment models.

\section{Conclusion}\label{sec:conclusion}

This letter showed that, with VMC on a PCC model, the behavior of a soft continuum arm is set by choosing and placing virtual elements, without planning, inverse kinematics, or a model of the environment. With the same hardware and controller, the arm shaped its whole body, held its tip near a point while keeping clear of a moving obstacle, moved along a chosen direction, rendered a tip stiffness, bounded its tip force, followed an unmodeled surface, reduced the force its body applied to an obstacle from $0.62~\text{N}$ to near zero, and avoided obstacles both autonomously and under teleoperation. The tip stiffness leveled off at about $5~\text{N/m}$: the stiffness reached in contact remains bounded by the compliance of the arm, keeping the interaction safe.

Regarding limitations, the encoder-based PCC shape estimate loses accuracy under contact, which embedded sensing could improve, and passivity holds only for fixed anchors and parameters, so stability conditions for variable impedance~\cite{kronanderStabilityConsiderationsVariable2016} would allow faster changes. Future work will investigate richer curvature models and the automatic tuning of virtual mechanisms.

\bibliographystyle{IEEEtran}
\bibliography{SoftArmControl}

\end{document}